\pdfoutput=1

\documentclass[11pt]{article}

\usepackage{times}
\usepackage{latexsym}

\usepackage[T1]{fontenc}

\usepackage[utf8]{inputenc}

\usepackage{microtype}

\usepackage{microtype}
\usepackage{graphicx}
\usepackage{subfigure}
\usepackage{booktabs} 
\usepackage{natbib}
\usepackage{graphicx}
\usepackage{amsmath}
\usepackage{amscd}
\usepackage{amssymb}
\usepackage{amsthm}
\usepackage{mathabx}
\usepackage{algorithmic}
\usepackage{algorithm}
\usepackage{booktabs}

\usepackage{multirow}
\usepackage{paralist}
\usepackage{color, colortbl}
\usepackage{enumitem}
\usepackage[normalem]{ulem}
\usepackage{stackengine}

\usepackage{xcolor}
\usepackage{tikz}
\usetikzlibrary{automata, arrows}
\usepackage{pgfplots}

\usetikzlibrary{arrows, positioning, calc}
\definecolor{color1}{RGB}{27,158,119}
\definecolor{color3}{RGB}{217,95,2}
\definecolor{color2}{RGB}{117,112,179}

\usetikzlibrary{shapes.geometric}
\newcommand{\En}{$\texttt{En}$}

\newcommand{\Gu}{$\texttt{Gu}$}

\newcommand{\Kk}{$\texttt{Kk}$}

\newcommand{\Tr}{$\texttt{Tr}$}

\newcommand{\X}{$\texttt{X}$}
\newcommand{\Y}{$\texttt{Y}$}

\newcommand{\minus}{\scalebox{0.7}[1.0]{$-$}}

\usepackage{naacl2021}

\usepackage{times}
\usepackage{latexsym}

\usepackage[T1]{fontenc}

\usepackage[utf8]{inputenc}

\usepackage{microtype}

\title{Towards Universality in Multilingual Text Rewriting}
\author{Xavier Garcia \and Noah Constant \and Mandy Guo \and Orhan Firat \\
         Google Research \\
         Mountain View \\
         California \\
         \texttt{\{xgarcia,nconstant,xyguo,orhanf\}@google.com}
         }
\begin{document}
\pgfplotsset{compat=1.6}
\maketitle
\begin{abstract}

In this work, we take the first steps towards building a \textit{universal rewriter}: a model capable of rewriting text in any language to exhibit a wide variety of attributes, including styles and languages, while preserving as much of the original semantics as possible. In addition to obtaining state-of-the-art results on unsupervised translation, we also demonstrate the ability to do zero-shot sentiment transfer in non-English languages using only English exemplars for sentiment. We then show that our model is able to modify multiple attributes at once, for example adjusting both language and sentiment jointly. Finally, we show that our model is capable of performing zero-shot formality-sensitive translation.

\end{abstract}

\section{Introduction}

Recent years have seen impressive developments in three areas of natural language processing: (1)~multilingual models \cite{lample2019cross,liu2020multilingual,xue2020mt5}, (2)~zero/few-shot text generation, either by large language models \cite{radford2019language,brown2020language} or more focused approaches \cite{xu2020variational,riley2020textsettr}, and (3)~controllable generation \cite{dathathri2019plug,chan2020cocon, keskar2019ctrl}. Despite the achievements in each of these areas individually, the space at their intersection remains understudied. Most\footnote{By few-shot generation, we refer to the ability to perform a variety of generative tasks given only a few examples. For this reason, we do not count zero-shot machine translation models (since those are restricted to only machine translation) or multilingual models performing zero-shot classification (which is not a generative task).} multilingual models have not explored zero/few-shot or controllable generation, and most text generation models (both zero/few-shot and controllable) have focused primarily on English and most recently code \cite{chen2021evaluating}.

\begin{figure}
\centering
\includegraphics[width=\columnwidth]{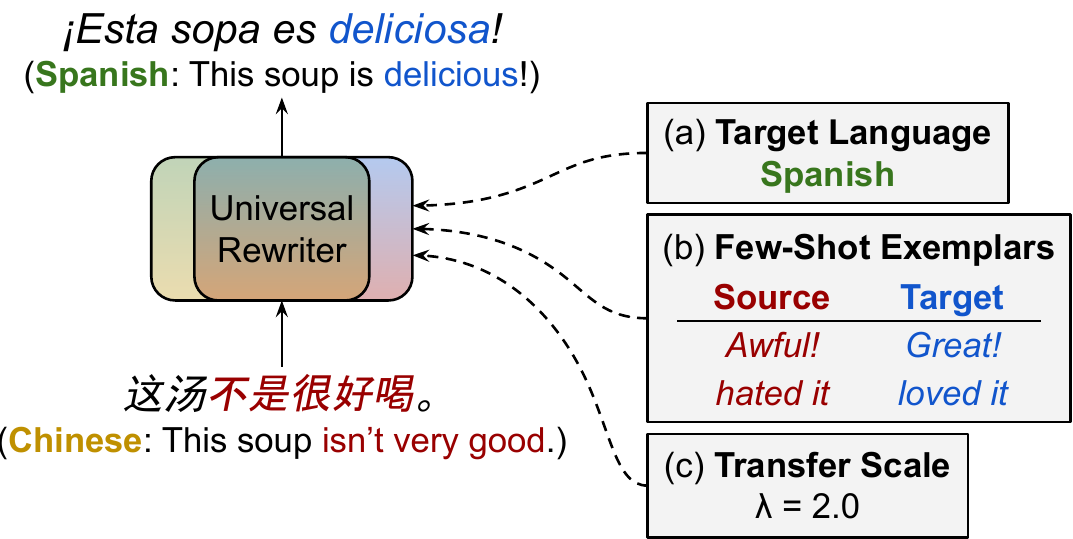}
\caption{\textbf{Our Universal Rewriter model can rewrite text to exhibit new attributes, such as changes in language, sentiment or formality.} (a) The target language is signaled by a unique language code. (b) Other attributes are controlled through few-shot examplars, which may leverage distinct languages from the input and target. (c) Transfer scale $\lambda$ modulates the strength of the attribute transfer.}
\label{fig:inference}
\end{figure}

We aim to bridge this gap by studying multilingual text generation in low-resource scenarios, which are one of the most natural settings to study zero/few-shot and controllable text generation. Some preliminary experiments in this direction were conducted by \citet{xue2020mt5}, who investigate the performance of mT5 (a large multilingual seq2seq model) on zero-shot multilingual classification tasks solved through generation. These experiments were limited to \textit{within-language} generation: generating output text in the same language as the input, which precludes many potential tasks of interest such as attribute-controlled machine translation e.g. formality-aware machine translation \cite{niu-etal-2017-study, niu-etal-2018-multi, wang2019harnessing}, adapting translation models to language varieties \cite{lakew-etal-2018-neural,kumar2021machine}, personalized machine translation \cite{michel2018extreme, vincent2021towards}, etc. When fine-tuning only on English data, the authors also encountered an unexpected phenomenon of ``accidental translation'' on the zero-shot tasks, where the model would provide the correct answer but in the wrong language. Moreover, although they only leveraged English data for the fine-tuning, the number of available examples for this task was in the thousands, which may be unrealistic for more restrictive generation tasks where finding numerous exemplars exhibiting a particular attribute may be difficult.

One such task is \textit{text style transfer}, which consists of transforming text from one style to another while preserving as much of the semantics as possible. Despite the ubiquity of stylized content across all languages, the vast majority of the text style transfer literature has focused exclusively on English. This is likely due to the scarcity of labeled data for many languages (known as low-resource languages). Given the success of zero-shot text generation observed with English models, we conjecture that large models trained on massive multilingual corpora should be capable of understanding textual attributes through a few exemplars in a given language and generalize to other languages in a zero-shot fashion.


In this paper, we carefully investigate this conjecture by leveraging a large pretrained multilingual model and fine-tuning it to perform general-purpose multilingual attribute transfer, extending the monolingual approach of \citet{riley2020textsettr}. We name the resulting model \textit{Universal Rewriter}, illustrated in Figure \ref{fig:inference}. Our main contributions are as follows: (1)~We showcase that few-shot approaches naturally extend to the multilingual setting. (2)~We show that multilingual models can endow text with additional attributes in one language using exemplars of that attribute from another language i.e.~\textit{zero-shot style transfer}. (3)~We demonstrate our model is capable of manipulating both language and textual attributes jointly. (4)~We demonstrate for the first time the possibility of zero-shot formality-sensitive translation, with no labeled data in the target language.

\section{Methodology}
\label{sec:background}

In this section, we first review TextSETTR \cite{riley2020textsettr}, an English-language few-shot style transfer model that serves as the starting point for our approach. Next, we discuss several design choices that are prohibitive when moving to the multilingual setting, and present our solutions to these problems.

\paragraph{Terminology} We use the terms \textit{attributes} and \textit{styles} interchangeably, though we emphasize that our model can go beyond the traditional tasks associated with style transfer, for example extending to translation (language transfer) and style-conditioned translation. We use the term \textit{exemplars} to denote inputs show-casing a particular attribute.

\subsection{Background: TextSETTR}

TextSETTR consists of a traditional encoder-decoder Transformer \cite{vaswani2017attention} architecture, equipped with an additional ``style extractor'' module which is structurally equivalent to the encoder component of the Transformer. The style extractor inputs exemplars of a style and extracts a fixed-width ``style vector'' by mean-pooling the output embeddings across tokens. These style vectors (and combinations thereof) can then be added element-wise to the encoded representations of a given input before being passed to the decoder in order to effect targeted style changes.



TextSETTR is trained in an unsupervised fashion, so it can handle arbitrary styles at inference time. To compensate for the lack of training labels, TextSETTR relies on the intuition that style is a ``slow-moving feature'', consistent across long spans of text \cite{akama-etal-2018-unsupervised,lample2018multiple}. Given a pair of consecutive sentences $(s_1, s_2)$ and a corruption function $\texttt{C}$, the model is trained to reconstruct $s_2$ given a corrupted variant $\texttt{C}(s_2)$ and the style of the previous sentence $s_1$. Here we think of $s_1$ as an exemplar of the style of $s_2$. These pairs of sentences are extracted using a heuristic sentence-splitter which splits on English punctuation.


TextSETTR employs two types of corruption: token-level and style-aware back-translation. In the former, a noise function is applied independently to each token, with a probability of either dropping a token, replacing it with one of the examples in the current batch at the same position or keeping it as is. In the latter, the model is used in inference mode to change the style of the input to match the style of a random sentence in the batch, and the resulting output is used as a corrupted version of the input.


\subsection{Moving to the multilingual setting}
\label{subsec:multiset}
We now outline where we diverge from TextSETTR as we enter the multilingual setting.

\paragraph{Architecture} In this work, we also embrace the encoder-decoder architecture with an attribute extraction module. However, unlike TextSETTR, we do not separate the attribute extraction module from the encoder-decoder architecture, and instead share all the weights between this module and the encoder. In order to distinguish the two, we prepend the text with a unique token when performing attribute extraction, and take the representation of this token from the encoder's output as the fixed-width vector representation, instead of mean-pooling all the representations. This allows us to leverage larger models without incurring excessive memory costs.


\paragraph{Sentence-splitting} In preliminary experiments, we reused TextSETTR's sentence-splitting algorithm (inherited from T5, and based on detecting English punctuation), but found this led to problems for some languages. In particular, this preprocessing step would discard any data which did not include ASCII punctuation. For languages like Chinese and Thai, which do not typically use ASCII punctuation, this resulted in filtering out most well-formed text, and leaving mainly text in the wrong language, or non-natural language (e.g.~Javascript code). We thus take a more language-agnostic approach and instead extract pairs of random non-overlapping spans of tokens from each line of text, and use the first-occurring span in the text as an exemplar of the attributes of the second span. This approach allows us to retain all data, is independent of language, and still exploits the ``slow-moving feature'' intuition used in the TextSETTR work.

\paragraph{Warm-starting} We start with an mT5 \cite{xue2020mt5} checkpoint as our initial model. In particular, we use the XL variant.

\paragraph{Corruption functions} We employ the same token-level noise used in TextSETTR\@. While we could also use style-aware back-translation as-is as done in previous work \cite{lample2018multiple,riley2020textsettr}, we take this opportunity to add an additional cross-lingual learning signal to our training objective by forcing the model to not only perform an attribute transfer procedure but also to translate the sentence to English. We choose English in particular due to the availability of parallel data between English and many languages. To make this objective more similar to the cycle-consistency found in traditional back-translation, we apply transfer using the negation of the ``true'' exemplar vector associated with the input (as opposed to leveraging a random exemplar). We perform the decoding operation using sampling, with a temperature of 1.5.

\paragraph{Translation data} In addition to our cross-lingual treatment of the style-aware back-translation objective, we also include translation as another cross-lingual learning objective by leveraging English-centric translation data. 

\paragraph{Inference strategy} There are multiple ways one could use the attribute vectors to perform rewriting. Suppose we are given sets of exemplars showcasing attributes $A$ and $B$, which we use to extract attribute vectors $V_A$ and $V_B$ respectively. Given an input $x$ with attribute $A$, we want to rewrite it to exhibit attribute $B$. TextSETTR's inference strategy is to first extract the attribute vector $V_x$ of $x$, then form the following attribute delta vector: $$V_{A,B,x} := V_x + \lambda (V_B - V_A)$$ where $\lambda$ is a scale factor, chosen by the user. The resulting vector $V_{A,B,x}$ then gets added to the encoded representation of $x$ before being passed to the decoder. For \textit{within-language} tasks, where the output should be in the same language as the input, we will use this inference strategy. In our early \textit{cross-language} experiments, however, we found the model was more reluctant to change languages when using this strategy. For this reason, we do not include the vector $V_x$ in the computation of $V_{A,B,x}$ for cross-language tasks, and instead use the following equation: $$V_{A,B,x} = \lambda (V_B - V_A).$$

\section{Experiments}
\label{sec:experiments}

In this section, we describe our main set of experiments. We first discuss our selection of languages and data sources considered, as well as data preprocessing steps. Next, we will measure the performance of our models in the challenging tasks of low-resource and unsupervised machine translation which we view as rewriting text in a different language. After this experiment, we introduce a multilingual variant of the sentiment transfer task typically explored in text style transfer literature which will allow us to explore the interaction between multilinguality and style transfer. Finally, we consider the problem of multiple attribute rewriting through two cross-lingual sentence rewriting tasks.

\paragraph{Data} We draw our monolingual data from mC4, the same dataset used to train mT5. We remove the ``\minus{}Latin'' languages due to their small data size. We leverage parallel data for 46 of these languages with English, by taking all language pairs with more than 500,000 lines of parallel data from the OPUS 100 \cite{zhang2020improving} dataset, with the exception of Hebrew, Croatian, Bosnian\footnote{These were excluded since their language codes in OPUS were not present in the language codes of mC4.}, and Japanese. We exclude Japanese as it appears in our sentiment evaluation, and we wish to test our model's ability to perform zero-shot attribute transfer in languages where no parallel data is available. As Japanese has a unique script and no genealogical relation with other languages under consideration, it poses a particularly challenging case.


\paragraph{Data sampling scheme}

We use both parallel and monolingual datasets for training our rewriter models. For each batch, we first randomly select to sample from either the monolingual or parallel datasets (with equal likelihood), and then uniformly sample from all available datasets in the chosen category.

\paragraph{Preprocessing} We process the monolingual data into input-exemplar pairs, using the approach described in \S\ref{subsec:multiset}. We additionally discard any pair where either element is shorter than five tokens. In the case of parallel data, we do not leverage any exemplars, and use a vector of zeros instead. For both data sources, we discard any training example with either inputs, exemplars or targets (in the case of parallel data) longer than 64 tokens. We use language-specific tokens pre-pended to the input to prime the model for translations. Instead of introducing new tokens, we re-use the tokens associated to the language names in English.

\paragraph{Training hyperparameters and settings}

The models were trained in JAX \cite{bradbury2018jax}. We use the Adafactor optimizer \cite{shazeer2018adafactor}, and reuse the accumulated optimizer states from the original mT5 training. We use a constant learning rate of 1e-3 and train for 25,000 steps, using batches of 512 inputs. 


\paragraph{Evaluation} For machine translation, we utilize BLEU scores \cite{papineni2002bleu}, computed through the sacreBLEU\footnote{case.mixed+numrefs.1+smooth.exp+tok.13a+version.1.3.0.} library \cite{post-2018-call}. For Gujarati, we follow standard practice and tokenize the output using the tokenizer provided by the Indic-NLP \cite{kunchukuttan2020indicnlp} library. For decoding, we use beam search, with a beam size of 5 for machine translation (\S\ref{subsec:mt}) and within-language sentiment transfer (\S\ref{subsec:zero-shot}), and beam size 1 for cross-language sentiment transfer (\S\ref{subsec:mult-attr}) and formality-sensitive translation (\S\ref{subsec:formality}).


\paragraph{Baselines} Given the absence of previous work on zero- or few-shot multilingual text rewriting, we introduce several new baselines to make up for the lack of prior baselines. These baselines follow a similar training recipe as our main rewriter model, but with some features removed. In addition to providing comparisons for the reader, we hope the inclusion of these baselines will justify our final training regimen. We describe the baselines below:

\begin{itemize}[itemsep=1pt]
    \item No parallel data. (\minus{}para)
    \item No language tokens. (\minus{}lang tokens)
    \item No exemplars in training. (\minus{}exemplars)\footnote{For this baseline, we also remove the back-translation step, to more closely mimic current state-of-the-art multilingual machine translation models \cite{siddhant-etal-2020-leveraging, garcia2020harnessing,liu2020multilingual} that leverage monolingual data.} 
    \item No back-translation. (\minus{}BT)
\end{itemize}

The ``\minus{}para'' setting tests whether the additional supervision signal obtained from the parallel data is necessary or even useful, and the ``\minus{}lang tokens'' setting addresses a similar question regarding language tokens. The ``\minus{}exemplars'' setting tests whether we suffer any degradation in translation quality by introducing the ability to leverage exemplars at inference time. Finally, given the expensive nature of back-translation, the ``\minus{}BT'' setting tests whether this objective is truly necessary. 
\subsection{Machine translation for low-resource languages}
\label{subsec:mt}

\begin{table*}
\small
\centering
\begin{tabular}{llccccccc}
\toprule
 & \textbf{Model} & \multicolumn{2}{l}{\stackanchor{\emph{newstest2019}}{\Gu{} \,$\leftrightarrow$\, \En{}\,\,}} &  \multicolumn{2}{l}{\stackanchor{\emph{newstest2018}}{\Tr{} \,$\leftrightarrow$\, \En{}\,\,}} &  \multicolumn{2}{l}{ \stackanchor{\emph{newstest2019}}{\Kk{} \,$\leftrightarrow$\, \En{}\,\,}}  \\ \midrule
\multirow{2}{*}{No parallel data}  & \textit{Universal Rewriter (\minus{}para)} & 0.5 & 0.3 & 2.1 & 2.0 & 0.6 & 1.4  \\ 
& \citet{kim2020and} & 0.6 & 0.6 & - & - & 0.8 & 2.0 \\ \midrule
\multirow{6}{*}{\stackanchor{No parallel data}{for \{\Gu{}, \Kk{}\}}}  &  \emph{Universal Rewriter}  & 9.6 & 23.2 & 16.3 & 23.7 & 5.8  & 19.8 \\
& \emph{Universal Rewriter (\minus{}lang tokens)}  & 0.0 & 17.8 & 10.7 & 22.3 & 0.6  & 17.8 \\
& \emph{Universal Rewriter (\minus{}exemplars)}  & 9.0 & 23.8 & 16.7 & 25.0 & 5.8  & 20.0 \\
& \emph{Universal Rewriter (\minus{}BT)}  & 8.8 & 23.0 & 16.5 & 24.0 & 5.3  & 19.9 \\
&  \citet{garcia2020harnessing}  &  4.4 & 19.3 &  8.4 & 15.9 & 3.9 & 14.8 \\
& mBART \cite{liu2020multilingual}  &  - & 13.8 & 17.8 & 22.5 &  - & -   \\  \midrule
\multirow{2}{*}{\stackanchor{With parallel data}{for \{\Gu{}, \Kk{}\}}} 
& \citet{garcia2020harnessing} & 15.5 & 19.3 & 18.1 & 22.0 & 9.5 & 15.1   \\
& mBART \cite{liu2020multilingual}  &  0.1 & 0.3 & 17.8 & 22.5 &  2.5 & 7.4  \\  \bottomrule
\end{tabular}
\caption{\textbf{BLEU scores of various supervised and unsupervised models for the low-resource language pairs.}  For any \X{}\,$\leftrightarrow$\,\Y{} language pair, the \X{}\,$\rightarrow$\,\Y{} translation results are listed under each \Y{} column, and vice-versa. There are multiple baselines available from \citet{garcia2020harnessing}, for fair comparison, we exclude the baselines which leverage offline back-translation. Note that the baseline without parallel data for \Gu{} and \Kk{} from \citet{garcia2020harnessing} also does not include parallel data for \Tr{}. }
\label{tab:benchmark-bleu}
\end{table*}

Our parallel data only accounts for a fraction of the languages available in mC4, yet we are interested in rewriting across all languages in mC4, potentially changing languages and attributes at the same time. To ensure our model is capable of such cross-lingual tasks, we inspect the translation quality on a selection of low-resource languages. We study both languages with parallel data as well as those without it, a setting referred to as ``unsupervised machine translation by language transfer'' or ``multilingual unsupervised machine translation'' \cite{garcia2020multilingual,garcia2020harnessing, liu2020multilingual, li2020reference}. In the latter setting, we are considering the translation quality for low-resource English-centric pairs where the associated low-resource languages have no parallel data at all, with English or otherwise. We select Turkish, Gujarati, and Kazakh as our low-resource languages and use the available \textit{newstest} test sets from WMT, using \textit{newstest2018} for Turkish and \textit{newstest2019} for Gujarati and Kazakh. Of these three languages, our model only sees parallel data for the language pair English-Turkish.

In Table~\ref{tab:benchmark-bleu}, we compare our Universal Rewriter with the bilingual unsupervised translation models of \citet{kim2020and}, as well as large multilingual unsupervised translation models \cite{garcia2020harnessing,liu2020multilingual} and their analogous supervised variants\footnote{To control the output language in our baseline with no language tokens, we leverage $2^{14}$ (16,384) examples of monolingual data for each language with $\lambda = 0.5$, chosen from \{0.5, 1.0, 1.5\} by looking at devset performance. This data was randomly drawn from the newscrawl datasets: \url{http://data.statmt.org/news-crawl/}.}. Notably, our rewriter attains comparable performance with the external baselines, despite not being directly trained for unsupervised translation. Interestingly, we find our baseline with no parallel data collapses, failing to produce the correct language and giving low BLEU scores. Such training instabilities have also been observed in other works; for instance, mBART \cite{liu2020multilingual} relied on a LangID filtering step during back-translation when training their unsupervised translation models. We see a similar pattern for our baseline without language tokens (\minus{}lang tokens) when translating into languages without parallel data. On the other hand, the remaining baselines which leverage parallel data manage to attain reasonable scores (even for language pairs without parallel data), justifying our inclusion of the parallel data. Finally, we observe that removing the ability to control text attributes via exemplars (\minus{}exemplars) confers no benefit for translation quality. This finding supports the view that a single general model can be well-suited to adjust both language and other textual attributes.


\subsection{Zero-Shot Sentiment Transfer}
\label{subsec:zero-shot}

Next, we turn to style transfer in a multilingual setting, and demonstrate for the first time the possibility of performing attribute transfer in one language using only exemplars from another language. Specifically, we observe that Universal Rewriter can transfer sentiment in French, German and Japanese (as shown in Table~\ref{tab:ex_sentiment}), despite having seen no sentiment labels in those languages, only four sentiment labels in English, and no labeled data whatsoever in Japanese.

Previous work on multilingual style transfer has been limited to a less ambitious setting where (i) models are trained for transferring one specific attribute, (ii) a large corpus of labels are provided at training time, and (iii) the labels cover all target languages \cite{briakou-etal-2021-ola}. By contrast, our approach is attribute-agnostic (one model can transfer arbitrary attributes), few-shot (only using labels at inference time), and zero-shot across languages.

As we are unable to use the XFORMAL \cite{briakou-etal-2021-ola} dataset due to legal restrictions, we develop our own multilingual style transfer evaluation, crafted from an existing dataset for multilingual classification. In particular, we leverage the multilingual Amazon reviews dataset provided by \citet{prettenhofer2010cross}. This dataset consists of Amazon reviews in English, French, German and Japanese with review ratings in the range 1--5. We treat reviews rated 4 or 5 as positive, and reviews rated 1 or 2 as negative, and drop reviews rated 3. We fine-tune mBERT, the multilingual variant of BERT \cite{devlin-etal-2019-bert}, on the remaining reviews and treat it as an oracle sentiment classifier\footnote{We split the reviews into training and development sets. This model achieved 80.1\% accuracy on this development set, which we considered sufficient for our purposes.}. We compare Universal Rewriter with our various ablation models on the task of transferring negative to positive sentiment.

\begin{table}
\centering
\includegraphics[width=\columnwidth, trim=0 15 0 0, clip]{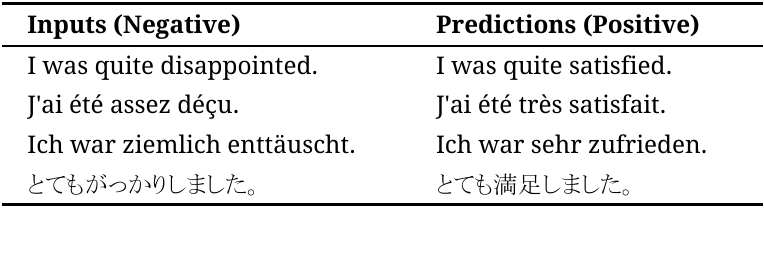}
\caption{\textbf{Examples of Universal Rewriter performing sentiment transfer in English, French, German and Japanese.} We use $\lambda = 5.0$, and define sentiment using just four English exemplars: \{ I loved it. / The movie was great. / I hated it. / The movie was awful. \} } 
\label{tab:ex_sentiment}
\end{table}

For evaluation, we remove examples longer than 61 tokens\footnote{We prepend to every example two tokens to control the language, and one end-of-sentence token. This gives us a maximum of 64 tokens, which is the maximum length that the models were trained on.}, and discard samples that disagree with our oracle classifier, only keeping those which it assigns above 80\% probability of being the correct class. We then construct development and test sets from the remaining examples, consisting of 800 examples, 200 from each language, equally balanced in positive and negative reviews.

For this experiment, we are interested in the following three questions:
(1)~Are English exemplars sufficient to perform sentiment transfer in non-English languages? (2)~Are we able to simultaneously transfer sentiment and language i.e.~multiple attribute rewriting? (3)~Is the use of parallel data necessary to achieve either of the above goals? To study the first question, we leverage twenty (total)  handcrafted English exemplars, exhibiting positive\footnote{\textbf{Positive exemplars:} I really enjoyed this movie. / They thought the food was very tasty. / It was perfect in every single way. / What a fantastic offer! / The most comfortable bed I've ever slept on, I highly recommend it. / I feel very happy. / I had a great time! /  You would have loved it! / A high quality item. / An amazing product!} and negative\footnote{\textbf{Negative exemplars:} I really disliked this movie / They thought the food was very bland. / It was horrible in every single way. / What a terrible offer! / The most uncomfortable bed I've ever slept on, I would never recommend it. / I feel very sad. / I had an unpleasant time. / You would have hated it. / A low quality item. / An awful product!} sentiment for all experiments. To address the third question, we purposely exclude English-Japanese parallel data. Given Japanese's unique script, we believe performance on Japanese sentiment transfer and translation is a reasonable proxy for the low-resource languages within mC4 for which we may have difficulties obtaining parallel data. Since model behavior depends on the inference-time parameter $\lambda$, we show results for a wide-array of $\lambda$ values (sweeping from 0.5 to 9.0 in increments of 0.5) to examine the influence of this parameter.


\begin{figure}
\centering
\includegraphics[scale=0.175]{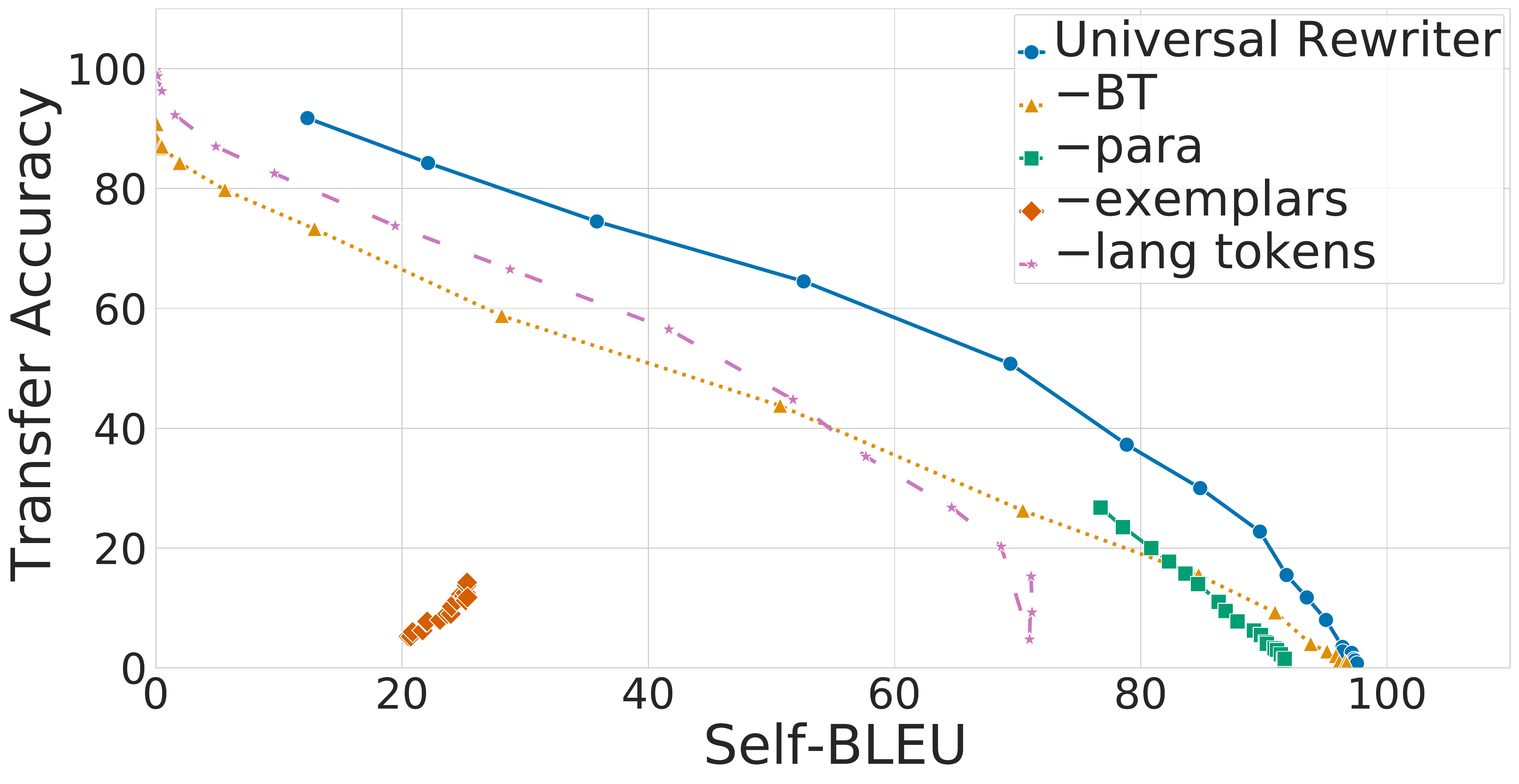}
\caption{\textbf{Measuring the trade-off between content preservation and sentiment transfer for the within-language sentiment transfer task.} We evaluate our rewriter models for $\lambda$ taking values in the set $\{0.5, 1.0, 1.5, ... , 8.5, 9.0\}$, using the same set of exemplars and models for each $\lambda$. }
\label{fig:within}
\end{figure}

For this task, we use the following two metrics: \textit{self-BLEU}, defined as the BLEU score using the inputs as references and the model's outputs as hypotheses, and \textit{transfer accuracy}, defined as the average sentiment transfer accuracy as judged by our classifier. We plot the results in Figure~\ref{fig:within}.

As expected, the $\lambda$ parameter allows us to control\footnote{Unsurprisingly, models trained without exemplars are unable to perform the transfer successfully.} the content preservation vs.~transfer accuracy trade-off, accurately changing attributes while still preserving content and language.  We note that the general shape of the baseline without language tokens (\minus{}lang tokens) is similar to Universal Rewriter, only shifted to the left. Upon inspection, we found this to be due to wrong-language errors, which negatively affected the self-BLEU score. 
\begin{figure*}[t!]
  \centering
  \subfigure[{\small Unsupervised language pairs}]{
\includegraphics[scale=0.23]{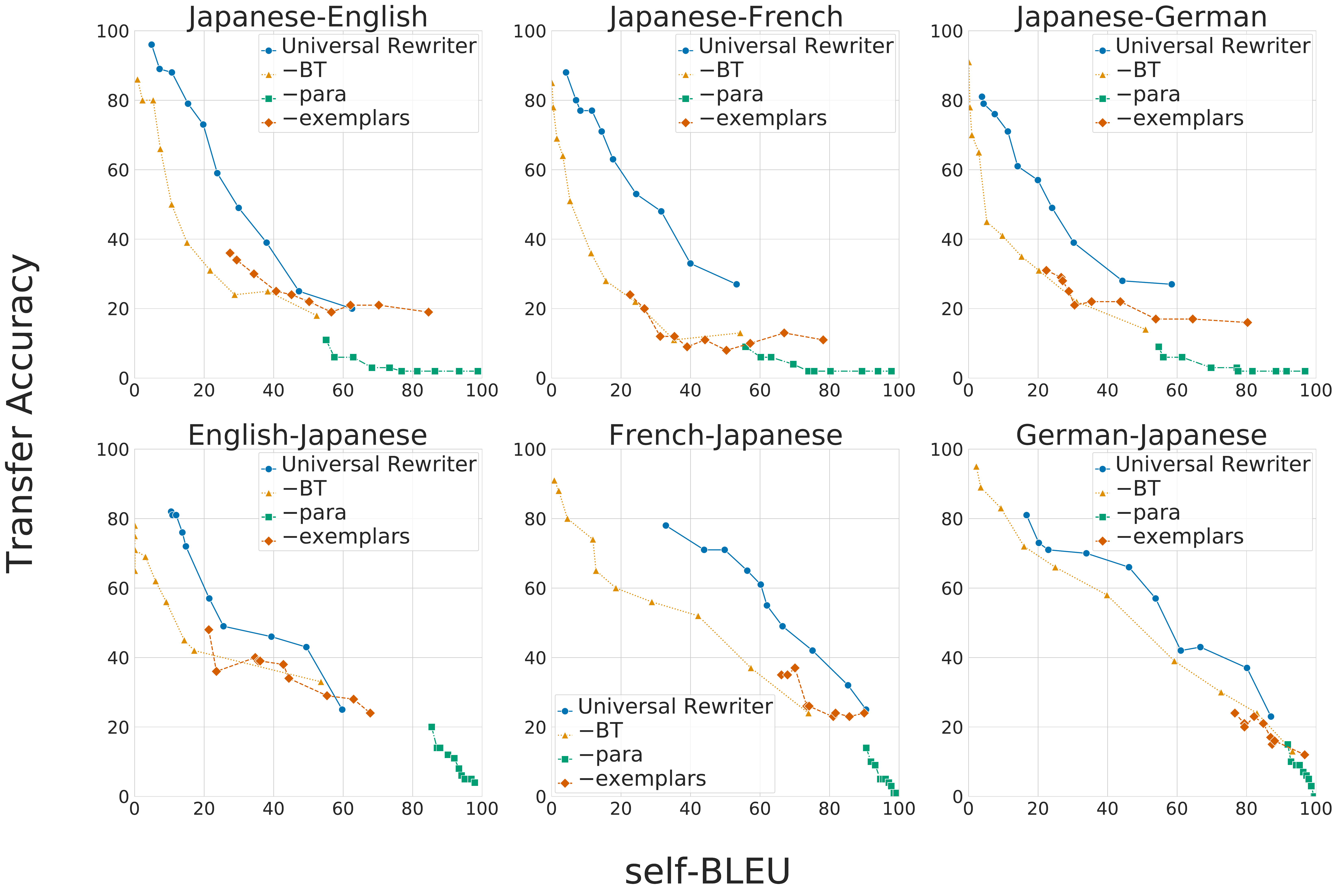}
\label{pic:supervised}
  }
 \\
  \subfigure[{\small Zero-shot language pairs}]{
\includegraphics[scale=0.23]{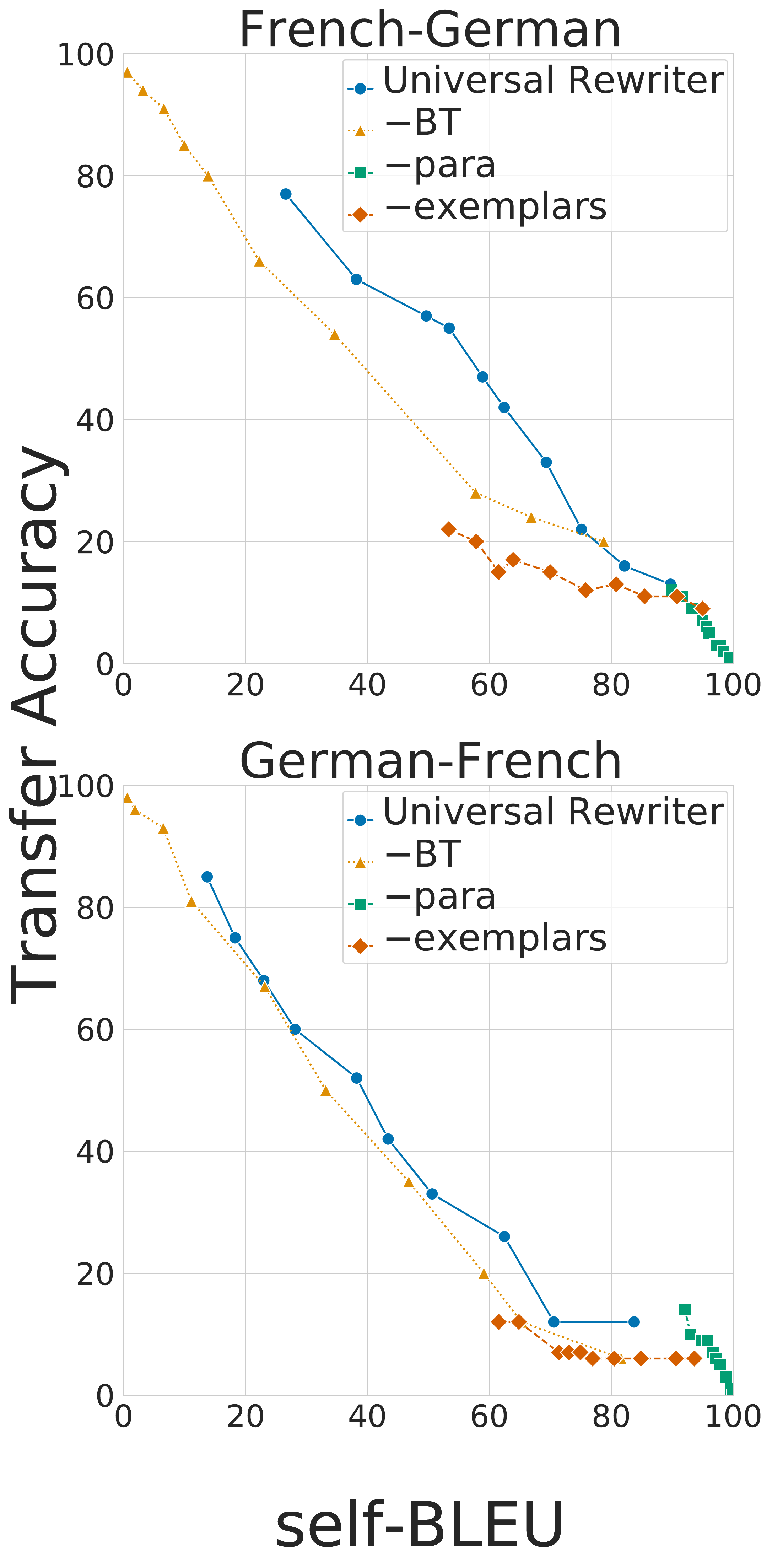}}
  \subfigure[{\small Supervised language pairs}]{
\includegraphics[scale=0.23]{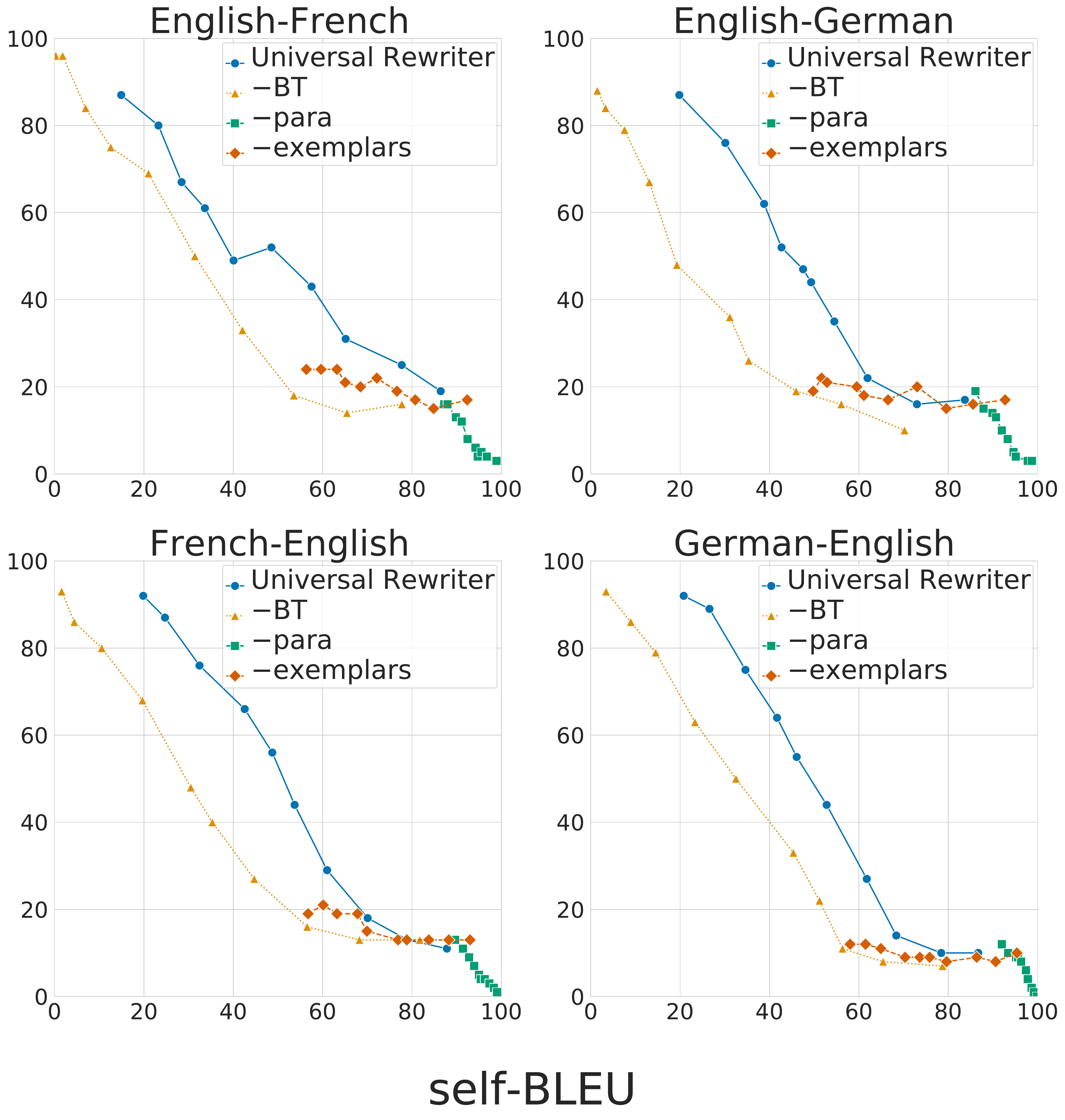}
  }
 \caption{\textbf{Measuring content preservation vs.~transfer accuracy as we change both language and sentiment at the same time.}  We evaluate our rewriter models for various $\lambda \in \{0.5, 1.0, 1.5, ... , 5.0\}$, using the same set of exemplars and models for each $\lambda$. The self-BLEU is computed with the translations generated with $\lambda = 0$. We group the results by whether the language pairs are (a) unsupervised, (b) zero-shot or (c) supervised.}
  \label{fig:cross}
\end{figure*}

\subsection{Cross-lingual Sentiment Transfer}
\label{subsec:mult-attr}

\begin{table*}[h!]
\centering
\includegraphics[width=\textwidth]{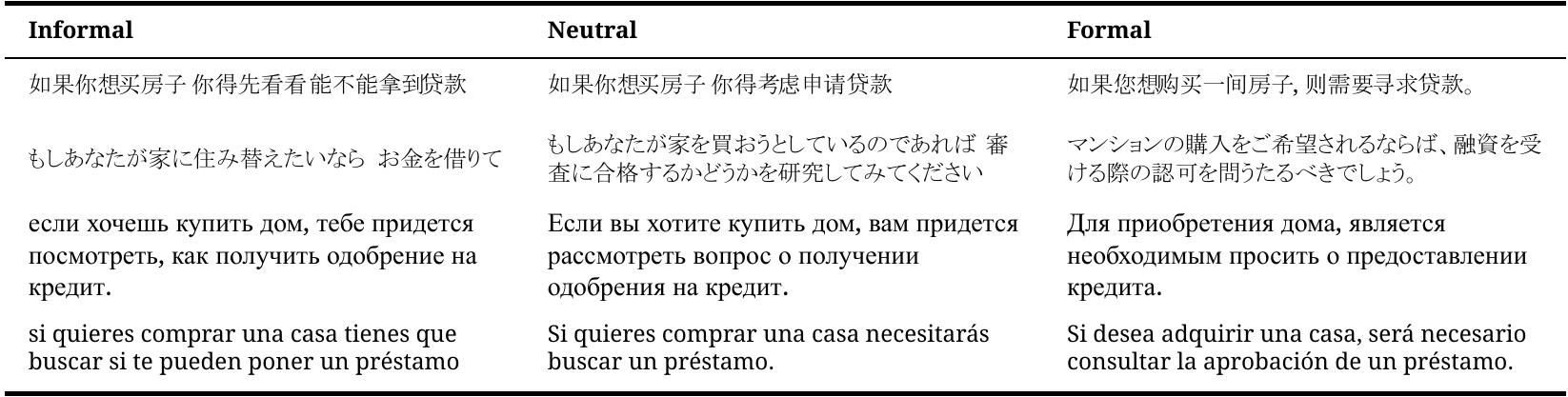}
\caption{\textbf{Formality-sensitive translations output by Universal Rewriter using just 10 English exemplars (total) to define formal and informal styles}. The target languages are Chinese, Japanese, Russian and Spanish, and the input sentence is ``If you want to purchase a house you'll need to look into getting approved for a loan.'' We use $\lambda = 1.7$ to achieve formal or informal style, and $\lambda = 0.0$ for neutral style.}
\label{tab:ex_formality}
\end{table*}

We investigate whether our model is capable of rewriting multiple attributes in one pass by assessing its ability to perform sentiment transfer and translation simultaneously. We re-use the same exemplars, development and test sets as our previous experiment, but instead of simply transferring sentiment, we also vary the target language to produce translations into each other language. In this cross-lingual setting, defining content preservation is more challenging, as self-BLEU is no longer a meaningful measure. Given the strong performance of our model on translation (\S\ref{subsec:mt}), we  generate ``neutral'' translations using $\lambda=0$ and treat these as gold labels. We then re-define self-BLEU to be the BLEU score computed using these predictions as references. As we increase $\lambda$, we expect to see more changes relative to the neutral translations, but also an increase in transfer accuracy. 

We consider three levels of supervision: (1)~\textit{unsupervised translation} for the language pairs Japanese $\leftrightarrow$ \{English, French, German\}, (2)~\textit{zero-shot translation} for the language pairs French $\leftrightarrow$ German, which lack parallel data but both languages have parallel data with English and (3)~\textit{supervised translation} for the language pairs \{French, German\} $\leftrightarrow$ English which have parallel data. Note that for the zero-shot (and some of the unsupervised) pairs, exemplars are in a different language from \emph{both} the input and output, which acts as an additional test of the cross-linguality of our model. For this task, we vary $\lambda$ between 0.5 and 5.0 with 0.5 increments. We omit the ``\minus{}lang tokens'' baseline since we have no explicit way of controlling both language and style for this baseline. As in the within-language case, our Universal Rewriter outperforms all our ablations, as seen in Figure~\ref{fig:cross}.


\subsection{Formality-Sensitive Translation}
\label{subsec:formality}

As a more practical test of our model's ability to transfer multiple attributes in one pass, we briefly explore formality-sensitive translation. Introduced by \citet{niu-etal-2017-study, niu-etal-2018-multi}, this task requires the model to translate input text into another language, while simultaneously controlling the formality level of the output text. This type of control is useful when the desired formality level is known ahead of time, and may not match that of the source text---for example, when providing translations for second language learners, or for use within an assistive agent.

As before, we assume labels are only available in English, and in limited quantity. Specifically, we construct 10 English exemplars (total) of formal\footnote{\textbf{Formal:} This was a remarkably thought-provoking read. / It is without question amongst my favorites. / We humbly request your presence at our gala in the coming week. / Loitering is strictly forbidden on the premises. / Were I to be approached for such a role, it would be the utmost honor.} and informal\footnote{\textbf{Informal:} reading this rly makes u think / Its def one of my all time favs / come swing by our bbq next week if ya can make it / you're not allowed to hang around here / that would be sooo dope if they pick me for the part} language.
Table~\ref{tab:ex_formality} shows a successful example of Universal Rewriter translating an input sentence into four languages at three levels of formality. Native speaker consultants confirm that, while not every translation is perfectly natural, they are all understandable, and the formality level clearly increases going from Informal to Neutral to Formal. Interestingly, these differences manifest in both lexical and grammatical choices, including the use of formal vs.~informal pronouns in Chinese, Russian and Spanish, despite this distinction being absent in English. This initial proof of concept demonstrates that our model is capable of zero-shot transfer of the notion of formality provided by the English exemplars onto other languages.

\section{Conclusion}

In this work, we undertook an initial exploration into creating a few-shot multilingual generative model, capable of rewriting text across a variety of languages, styles, attributes, etc.  We show that multilingual models, leveraging only a few exemplars in English showcasing an attribute, are able to rewrite text to exhibit this attribute across languages that they know.

\bibliography{anthology,custom}
\bibliographystyle{acl_natbib}
\end{document}